\pdfoutput=1
\documentclass[11pt]{article}

\usepackage[final]{acl}

\usepackage{times}
\usepackage{latexsym}
\usepackage{enumitem}
\usepackage{booktabs}   % For \toprule, \midrule, \bottomrule
\usepackage{array}      % For p{width} column types
\usepackage{caption}    % (Optional) For better control over table captions
\usepackage{alphabeta}
\usepackage{amssymb}
\newlist{greekenum}{enumerate}{1}
\setlist[greekenum,1]{label=\greek*.}

\usepackage[T1]{fontenc}
\usepackage[utf8]{inputenc}

\usepackage{microtype}

\usepackage{inconsolata}

\usepackage{graphicx}

\title{Patient-Centered RAG for Oncology Visit Aid \\Following the Ottawa Decision Guide
}

\author{
 Siyang Liu$^{1}$,
 \textbf{Lawrence Chin-I An\textsuperscript{2}},
 \textbf{Rada Mihalcea$^{1}$}
\\
 \textsuperscript{1}The LIT Group, Department of Computer Science and Engineering, \\University of Michigan, Ann Arbor,
 \\
 \textsuperscript{2} Rogel Cancer Center, University of Michigan, Ann Arbor,
\\
 \small{
   \href{mailto:lsiyang@umich.edu}{lsiyang@umich.edu},\href{mailto:mihalcea@umich.edu}{mihalcea@umich.edu}
 }
}

\begin{document}
\maketitle
\begin{abstract}
Effective communication is essential in cancer care, yet patients often face challenges in preparing for complex medical visits. We present an interactive, Retrieval-augmented Generation-assisted system that helps patients progress from uninformed to visit-ready. 
Our system adapts the Ottawa Personal Decision Guide into a dynamic retrieval-augmented generation workflow, helping users bridge knowledge gaps, clarify personal values and generate useful questions for their upcoming visits.
Focusing on localized prostate cancer, we conduct a user study with patients and a clinical expert. Results show high system usability (UMUX Mean = 6.0 out of 7), strong relevance of generated content (Mean = 6.7 out of 7), minimal need for edits, and high clinical faithfulness (Mean = 6.82 out of 7). This work demonstrates the potential of combining patient-centered design with language models to enhance clinical preparation in oncology care.
\end{abstract}

\section{Introduction}
People with cancer often face significant communication challenges throughout their healthcare journey. These challenges can begin at the time of initial diagnosis and continue through their experience of treatment, recovery, and survivorship. \cite{chen2024access, committee2001crossing}.
Strong bi-directional patient-provider communication is essential for the provision of patient-centered cancer care. However, the volume and complexity of new information that patients need to comprehend, underlying emotional distress and worry related to receiving a cancer diagnosis, and limited provider appointment times can make it difficult to establish an ideal healing relationship~\cite{epstein2007patient, mcginnis2015transforming}.
%  \cite{chen2024access, mcginnis2015transforming}.

Cancer is inherently complex due to its diverse subtypes and variable progression, genetic variations, risk factors (e.g., family history and race), and multiple treatment options \cite{litwin2017diagnosis}.
To make informed medical decisions, patients must process extensive information and reconcile it with their personal preferences, all within a limited consultation window~\cite{hamilton2017good}.
For example, patients diagnosed with localized prostate cancer (a type with relatively lower spread risk) often face at least three treatment options (e.g. active surveillance, surgery, radiation therapy), each associated with varying treatment experience and side effects affecting sexual, urinary, and bowel functions.
Moreover, patients bring individual backgrounds—such as race, interpersonal environment, and financial circumstances—and personal concerns, such as fear of sexual dysfunction, all of which further complicate the decision-making process~\cite{martinez2019shared}.

To support patients in preparing for these complex visits, we build a system that helps individuals with cancer transition from being uninformed to visit-ready.
Specifically, the system adapts the Ottawa Personal Decision Guide—traditionally a static worksheet—into an interactive, AI-assisted workflow powered by retrieval-augmented generation (RAG). 
RAG is used in two stages: surfacing the decision landscape and generating personalized visit questions. 
Our evaluation focuses on localized prostate cancer as a case of a multifaceted illness with numerous decision-making points compressed into relatively short periods of time.
Results show high overall usability and positive component-level feedback, indicating that the system effectively helps patients understand their decision landscape, clarify personal values, and prepare meaningful questions for their clinical visits.

\begin{figure*}[t]
    \centering
    \includegraphics[width=0.85\linewidth]{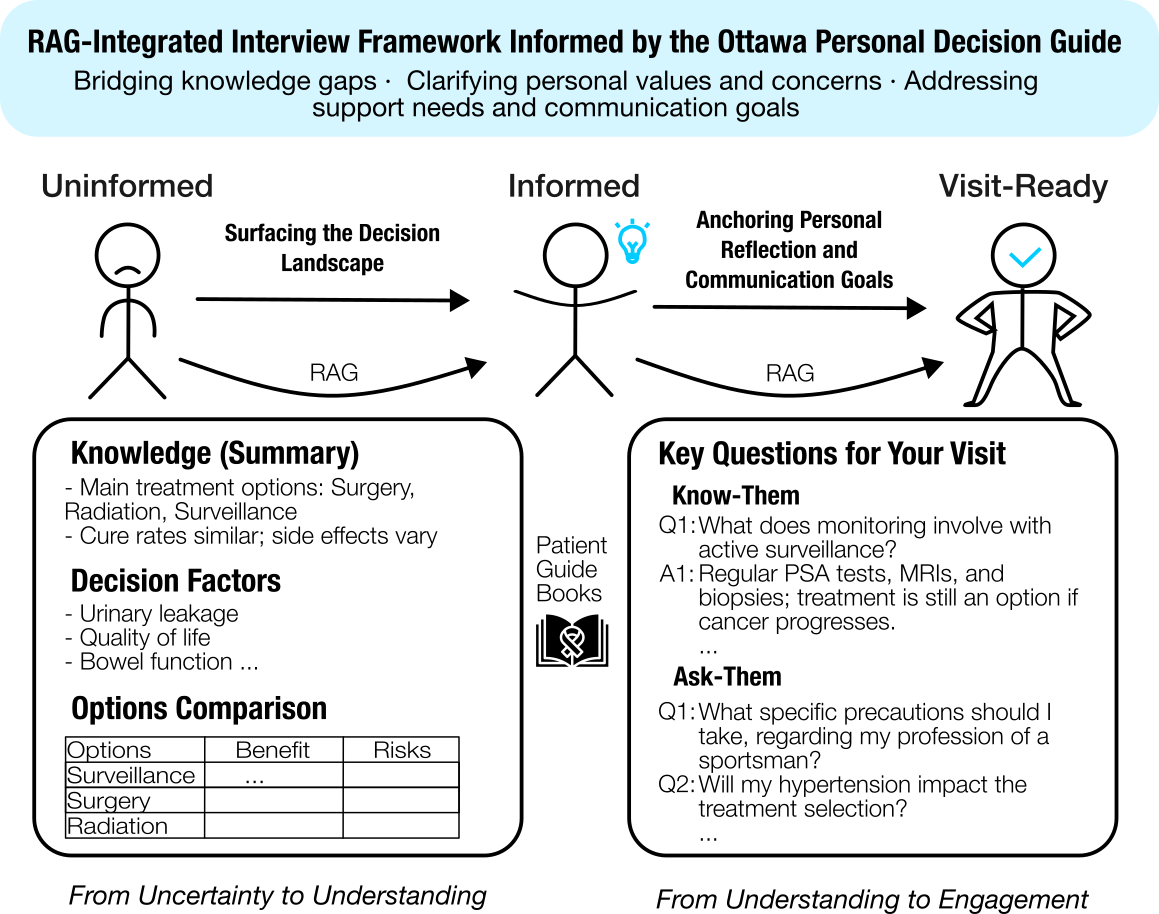}
    \caption{Design Ideas of Integrating RAG with the Ottawa Guide}
    \label{fig:main}
\end{figure*}

\begin{figure*}[t]
    \centering
    \includegraphics[width=\linewidth]{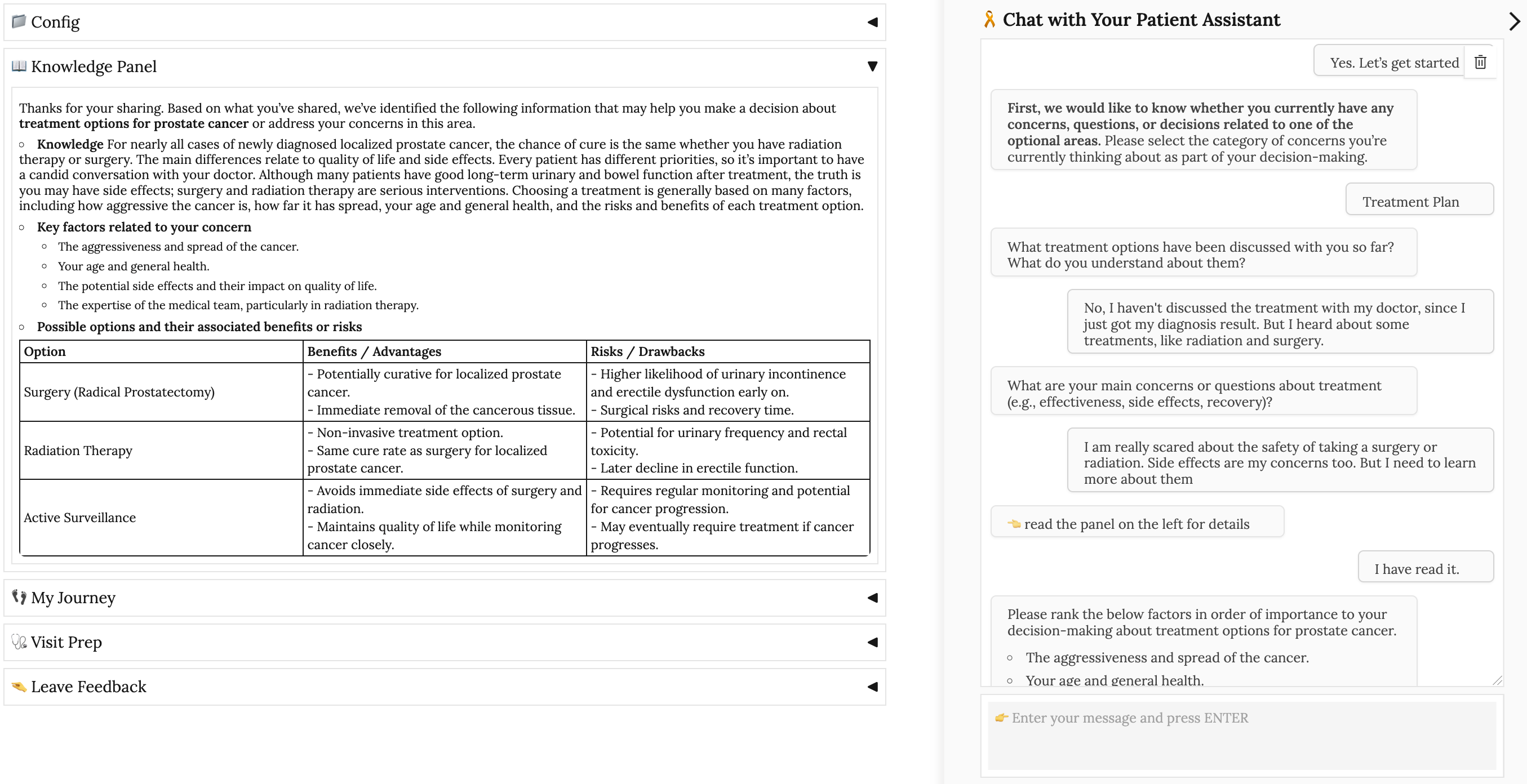}
    \caption{Interview Interface and Knowledge Panel Section}
    \label{fig:int_know}
\end{figure*}

\section{System Overview}

Our system is a web-based interactive tool designed to help patients prepare for oncology visits. You can try the demo at \url{https://cancervisitprep.webanonymous75.win}. 
The design integrates the Ottawa Personal Decision Guide (OPDG) ~\cite{oconnor2003dsf} with retrieval-augmented generation to bridge knowledge gaps, clarify personal values, and generate personalized questions for the upcoming medical visit.
This section presents two layers: the overarching design philosophy (Section~\ref{sec:design}) and the concrete interface components of the system (Sections~\ref{sec:interview}, \ref{sec:narrative}, and \ref{sec:output}).

\subsection{Design Ideas: Integrating RAG with the Ottawa Guide\label{sec:design}}

As shown in Figure~\ref{fig:main}, our framework builds on the Ottawa Personal Decision Guide~\cite{oconnor2003dsf}  but adapts it into an interactive, AI-supported experience.
The OPDG outlines three steps to help patients make informed, value-aligned decisions:
\begin{enumerate}[label=\roman*., leftmargin=*, labelindent=2em, itemsep=0em, parsep=0em]
  \item[\textsc{Step A}] Consider what you already know about the decision;
  \item[\textsc{Step B}]  Clarify personal values and preferences by reviewing options along with their risks, benefits, and levels of certainty;
  \item[\textsc{Step C}]  Identify support needs and plan the next steps.
\end{enumerate}

Traditionally, the OPDG is delivered as a worksheet or static form \footnote{OPDG worksheet: \url{https://decisionaid.ohri.ca/docs/das/OPDG.pdf}}.
However, many patients begin their decision-making with little background knowledge, which can make it difficult to meaningfully engage with the traditional format.

To address this gap, our system integrates retrieval-augmented generation into the OPDG process:
\begin{enumerate}
\item \textbf{Surfacing the Decision Landscape:} 
To support the transition from \textsc{Step A} to \textsc{Step B}, RAG addresses knowledge gaps by retrieving and summarizing background information, key decision factors, and structured comparisons of available options drawn from trusted cancer care guidebooks. This helps patients understand the scope of the decision they face.

\item \textbf{Anchoring Personal Reflection and Communication Goals:} Once patients have built a foundational understanding, they are guided to reflect on their personal priorities. To support the transition from \textsc{Step B} to \textsc{Step C}, RAG is then used to generate two types of visit questions, as exemplified in Figure~\ref{fig:main}.

\end{enumerate}

Together, these two stages support the patient's journey from uncertainty to understanding and from understanding to engagement.

\subsection{Structured Interview Interface\label{sec:interview}}
A structured interview is built for patients to chat with the system assistant. As shown in Figure~\ref{fig:int_know}, the right side of the screen features a chat assistant that guides patients step-by-step.
The interview starts by asking patients if they have any concerns, questions, or decisions to make. They can choose from broad topics including
\textit{Diagnosis and Screening}, \textit{Treatment Plan}, \textit{Physical Wellness}, \textit{Emotional and Mental Health}, \textit{Nutrition and Dietary Guidelines}, \textit{Long-Term Management and Monitoring}, \textit{Insurance and Financial Support}, or \textit{Other Concerns}.
The system also asks open-ended questions to learn more about what the patient knows or feels unsure about.

Based on their responses, the system identifies what the patient may not know and shows helpful information in the left panel (called the “Knowledge Panel”). This includes background facts, key points to consider, and a comparison of options. This step matches the first use of RAG in the system (explained in Section~\ref{sec:design}).

Once patients review the knowledge, the system moves to the next step—reflection. It asks questions like:
\textit{"Which factors are most important to you?"},
\textit{"Where are you still unsure?"}.
These questions help patients think clearly about their values and what matters most before talking to their doctor. This part supports the second RAG stage.

After finishing this reflection, patients click the \textit{“Generate My Journey”} button. This takes them to the next stage, where their personal summary is created (see Section~\ref{sec:narrative}).

\subsection{Editable Personal Narrative\label{sec:narrative}}
After completing the interview, the system generates a personalized, first-person summary that reflects the patient's current concerns, decision context, and values.
As shown in Figure~\ref{fig:int_journey}, the narrative is fully editable. 
Patients are encouraged to revise the text to ensure it accurately captures their experience.
This stage supports narrative agency, allowing patients to take ownership of their stories. It also ensures that the final visit preparation reflects the patient's authentic voice and priorities.

\subsection{Visit Preparation Output\label{sec:output}}

\begin{figure}[ht]
    \centering
    \includegraphics[width=\linewidth]{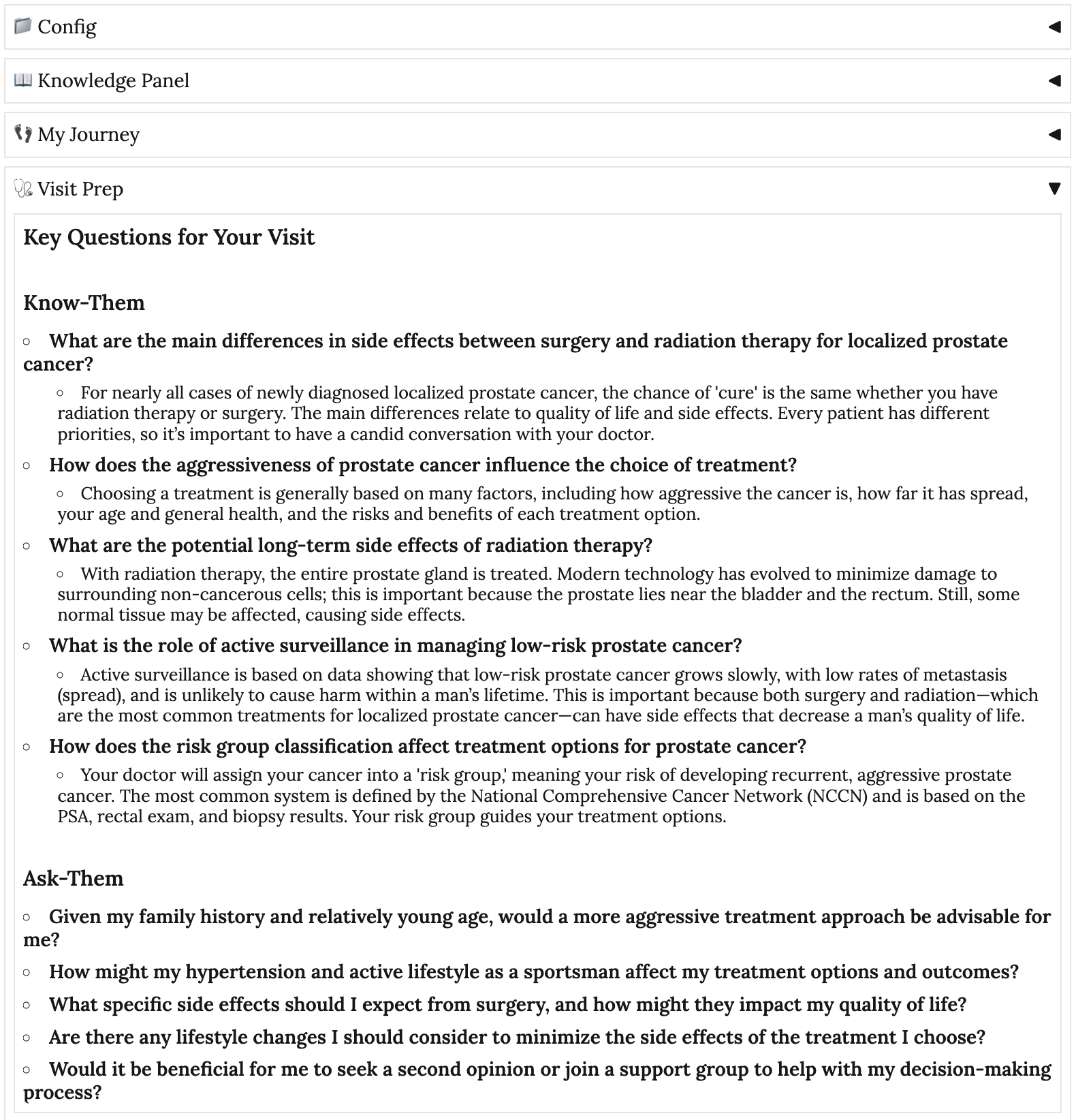}
    \caption{Visit Preparation Output Section}
    \label{fig:int_ques}
\end{figure}

Upon confirmation of the narrative, the second RAG module generates two sets of visit-prep questions, demonstrated in Figure~\ref{fig:int_ques}:
\begin{itemize}[ leftmargin=*, labelindent=0em, itemsep=0em, parsep=0em]
\item Know-Them Questions: Questions the patient should know the answer to, paired with short answers retrieved from validated patient educational guidebooks.
\item Ask-Them Questions: Personalized questions that the patient should raise with their doctor. These are questions that cannot be answered directly based on the contents of validated guidebooks and thus potentially high-value to discuss during the patient's upcoming visit. 
\end{itemize}
This final step supports confident, personalized communication during the clinical visit.

\subsection{Custom Book Uploader (Admins)}
\begin{figure}[ht]
    \centering
    \includegraphics[width=\linewidth]{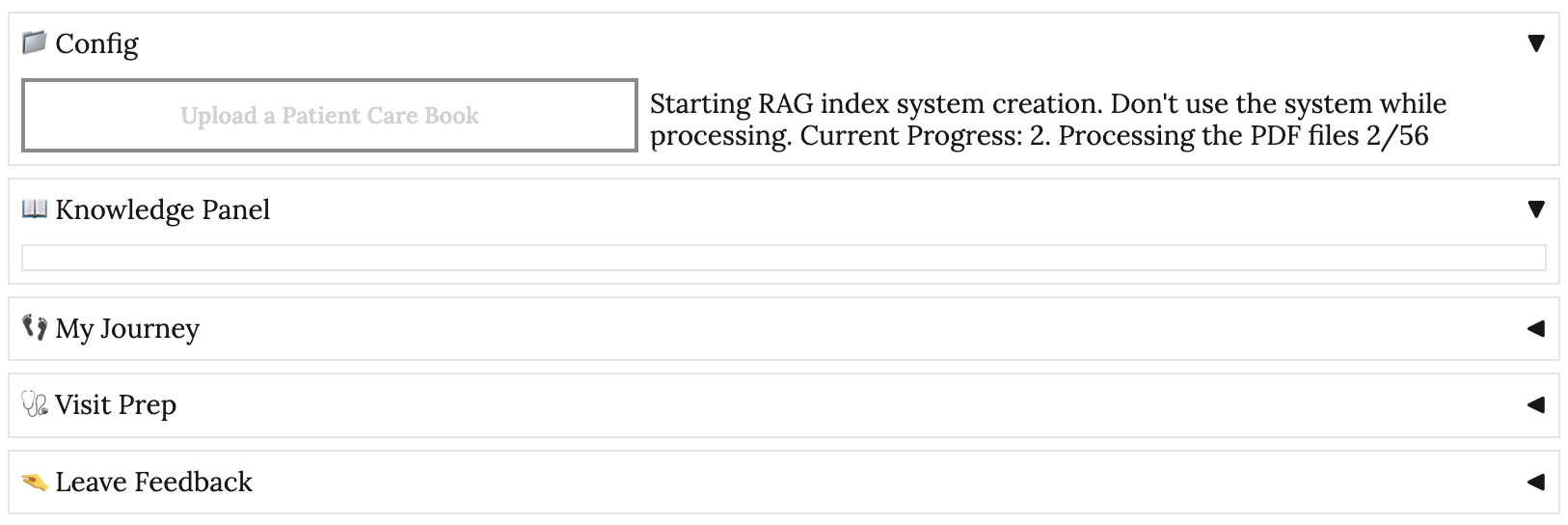}
    \caption{Custom Book Uploader Section}
    \label{fig:int_uploader}
\end{figure}

The system includes an admin-facing uploader (Figure~\ref{fig:int_uploader}) that allows uploading new cancer care guidebooks as backend sources, which supports the system adapted to different cancer types and existing resources.
Users can upload a folder containing all pages of a guidebook in PDF format. 
Then, the system automatically breaks the content into meaningful text segments, creates vector embeddings for each segment, and builds an index for semantic searching. 
A progress bar shows how far along the indexing process is.
This uploader makes it possible to flexibly support different use cases without needing to change the code.

\section{Implementation Details}

Our system is implemented in Python and deployed as a web-based application. The overall architecture consists of a frontend interface, a retrieval-augmented generation backend, and language model integration for generation and reasoning tasks.

The front end is developed using the Gradio framework\footnote{\url{https://gradio.app/}}, which supports the rapid development of web-based GUIs for demo machine learning models. It manages all patient-facing components, including the chat interface, knowledge panel, editable narrative view, and question summary board.

The backend is powered by a RAG architecture, built using LlamaIndex\footnote{\url{https://www.llamaindex.ai/}}. This includes: (1) embedding expert-verified cancer care guidebooks and storing them in a vector index for semantic search; (2) retrieving relevant knowledge segments using a similarity-based retriever; and (3) synthesizing responses by passing the retrieved segments to language models in response to task queries. 
These components are connected through a query engine that manages both retrieval and generation tasks.

We use OpenAI’s API~\cite{openai2023gpt4} to support language generation throughout the pipeline. The LLM is employed to generate knowledge panel content during the interview stage (decision landscape), rephrase interview transcripts into editable personal narratives, and generate personalized visit questions tailored to the patient’s needs. LLM-based embeddings are also used to construct and maintain the semantic index that supports information retrieval.

The system is hosted through a secure Cloudflare tunnel~\cite{cloudflared}, which creates encrypted access for secure communication between the user and the server. All user activities are logged and stored securely.

\section{System Evaluation}
% We conduct a component-wise evaluation of our system to assess the quality and usefulness of the generated content, as well as an overall evaluation of system usability.

We evaluate both the overall usability and individual components of our system to understand its effectiveness, usefulness, and limitations from the patient’s perspective, as well as its clinical faithfulness from expert review.

\subsection{Evaluation Overview}

The system is assessed through two complementary approaches:

\begin{itemize}[leftmargin=1em, itemsep=0em, parsep=0em]
    \item \textbf{System Usability Evaluation:} We use the UMUX (Usability Metric for User Experience) questionnaire~\cite{10.1016/j.intcom.2010.04.004} to measure user perceptions of effectiveness, satisfaction, efficiency, and overall on a 7-point Likert scale.
    
    \item \textbf{Component-wise Evaluation:} Each system module is evaluated based on its intended function and impact on the patient experience. These include the Knowledge Panel, Medical Journey generation, Visit Question generation, and the Faithful Use of Retrieved Information.

    We evaluate each core module of the system according to its function and impact on the user experience:
    \begin{itemize}[leftmargin=1.5em, itemsep=0em, parsep=0em]
        \item \textit{Knowledge Panel Materials:} Patients rate how relevant the knowledge summary, decision factors, and option grid are to their concerns on a 7-point scale.
        \item \textit{Medical Journey Generation:} We track how much users edit the journey text (e.g., token-level rewrites) as a proxy for how well the generation captures their intended meaning.
        \item \textit{Visit Question Generation:} Users indicate which of the five “know-them” questions improve their understanding, and which of the five “ask-them” questions they would consider bringing to a doctor.
        \item \textit{Faithful Use of Retrieved Information:} A clinical expert reviews whether the generated content accurately reflects the retrieved source material, or whether the generation presents medically sound reasoning based on it. Ratings are given on a 7-point scale.
    \end{itemize}
\end{itemize}

\subsection{Procedure and Participants}

To ensure the evaluation was focused and feasible, we limited the study to patients with localized prostate cancer—an early-stage form of the disease with low spread risk. A cancer expert on our team recommended the book \textit{Patient Guide to Localized Prostate Cancer}~\cite{patientguide2023} as the knowledge base for content generation.
The study is approved by the Institutional Review Board (IRB) at our university.

\paragraph{Patient study.} We recruited 10 male participants with experience of localized prostate cancer via Prolific\footnote{\url{https://www.prolific.com}}. Ages ranged from 39 to 66 (mean = 53.3, SD = 9.5). Two participants identified as White or Caucasian and eight as Black or African American.
Participants completed a consent form and a short demographic survey, then used the system remotely on their own devices. Before starting, they are guided to reflect on a moment during their cancer experience when they felt uncertain or overwhelmed about a medical decision. They are asked to imagine preparing for an upcoming visit and wanting to feel more confident in discussing their concerns with their doctor.=
After using the system, they fill out a feedback survey designed to evaluate the overall usability and each system component based on the criteria described above. 
The full session took about 35 minutes. 
Participants are compensated at a rate of \$14 per hour.

\paragraph{Expert evaluation.} One of the authors—a clinician affiliated with our university’s Cancer Center—to review whether the generated content accurately reflects the retrieved source materials.

\begin{figure}[t]
    \centering
    \includegraphics[width=\linewidth]{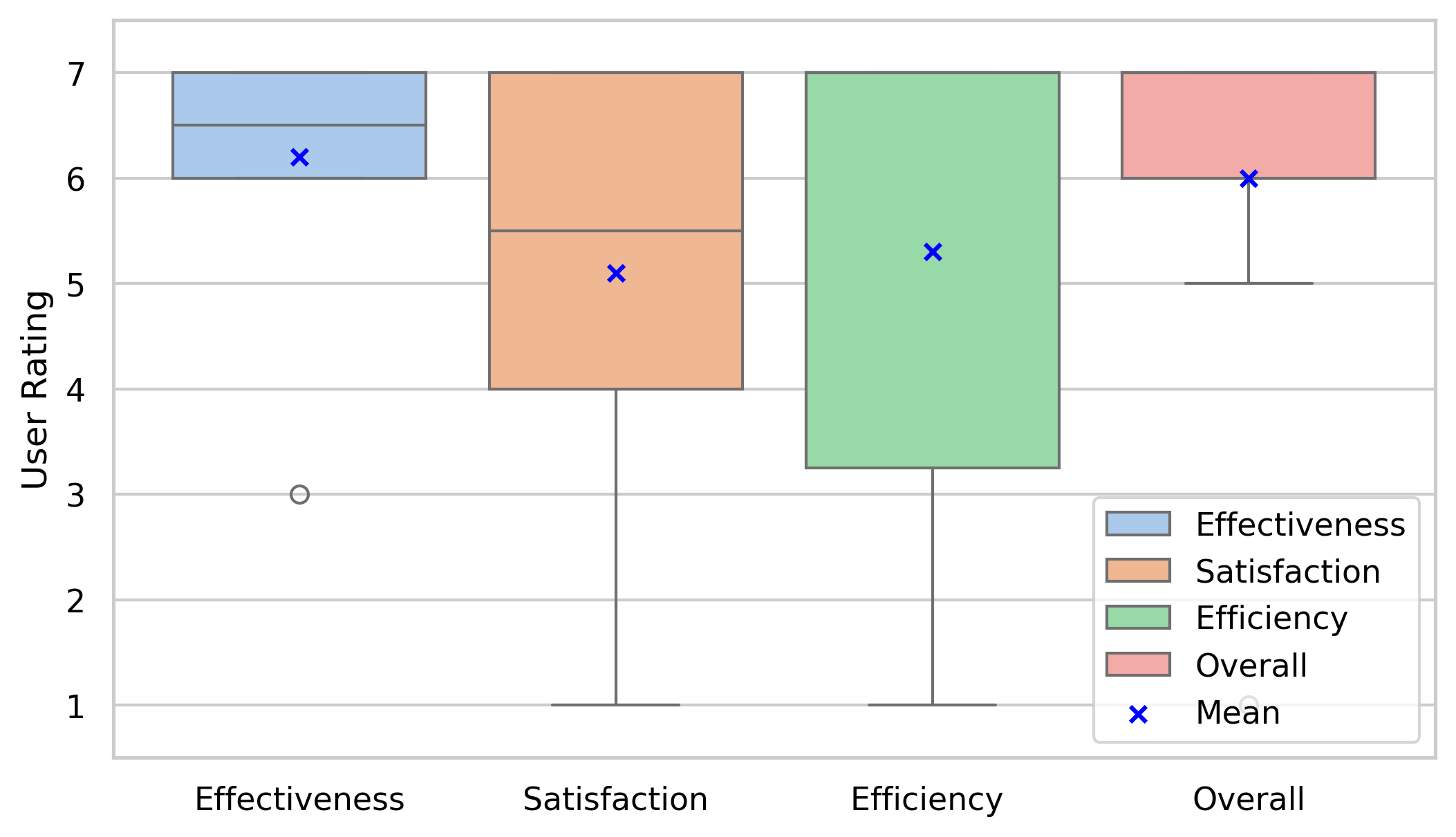}
    \caption{User ratings on system usability across UMUX dimensions. Scores are based on a 7-point Likert scale (1 = strongly disagree, 7 = strongly agree).}
    \label{fig:use_stats}
\end{figure}

\begin{table*}[t]
\centering
\begin{tabular}{llccc}
\toprule
\textbf{Component} & \textbf{Evaluation Metric} & \textbf{Mean} & \textbf{SD} & \textbf{N} \\
\midrule
\textbf{Knowledge Panel} & Relevance (1–7) & 6.7 & 0.48 & 10 \\

\textbf{Medical Journey Generation} & \% Participants Who Edited & 20\% & -- & 10 \\
& \% Token Changes (for edited cases) & 16\% & -- & 2 \\

\textbf{Visit Question Generation} & Helpful “Know-them” Count (0–5) & 3.3 & 0.82 & 10 \\
& Useful “Ask-them” Count (0–5) & 3.5 & 0.71 & 10 \\

\textbf{Faithful Use of Retrieved Info} & Expert Rating (1–7) & 6.82 & 0.40 & 10 \\
\bottomrule
\end{tabular}
\caption{Component-wise evaluation results. Ratings reflect subjective user scores, observed editing behavior, and expert assessments.}
\label{tab:component_eval}
\end{table*}

\subsection{Evaluation Results}

\subsubsection{System Usability}
As shown in Figure~\ref{fig:use_stats}, the system receives a high overall usability rating (Mean = 6.0). The effectiveness score is the highest, indicating that users find the system useful for its intended purpose. However, satisfaction (Mean = 5.1) and efficiency (Mean = 5.3), while still above neutral, are rated lower and with high variance. 
This appears somewhat inconsistent with the high overall score. To better understand this, we collect follow-up feedback from participants and identify two external influencing scores:

% \noindent\textbf{\textit{External factors:}}
\begin{itemize}[leftmargin=1em, itemsep=0em, parsep=0em]
    \item \textit{Emotion bias:} The satisfaction item in UMUX is negatively phrased: \textit{“Using this system is a frustrating experience.”} The satisfaction score is calculated by subtracting the agreement rating from 7. Participants explain that the emotional frustration comes from the frustrating nature of the illness, not the system itself. One participant shared:
    \begin{quote}
        The system itself means well, but having to go through it is very frustrating.
    \end{quote}

    \item \textit{Internet speed:} Some international users experienced slow system response, as they accessed it from other continents such as Africa. In real-world clinical settings, this issue would likely be minimized with a stable hosting environment.
\end{itemize}

% \noindent \textbf{\textit{Internal factors:}}
% \begin{itemize}[leftmargin=1em, itemsep=0em, parsep=0em]
%     \item \textit{Layout complexity:} A very few users found the interface unintuitive or difficult to navigate in certain parts. For example:
%     \begin{quote}
%         I had to go back and forth between different pages to complete a single task, which disrupted the flow. Some labels or button names weren’t immediately clear.
%     \end{quote}

%     \item \textit{Lack of personalization when the interview is sparse:} When users provided minimal input during the interview, the generated content felt generic. One noted:
%     \begin{quote}
%         The tool felt a bit basic and not super personal. It worked fine, just didn’t fully meet my needs.
%     \end{quote}
%     This suggests a need to balance the simplicity of the interview flow with the depth needed to explore patients’ experiences meaningfully.
% \end{itemize}

\subsubsection{Component-wise Evaluation Results}
The component-level evaluation results demonstrate that the system effectively delivers useful content to support patient understanding and preparation. As shown in Table~\ref{tab:component_eval}, participants strongly agree that the Knowledge Panel content was relevant to their concerns (Mean = 6.7). 
Participants also find the generated materials useful with minimal need for revision; only 2 out of 10 participants make any edits to the automatically generated personal journey text, suggesting that most users find the system’s paraphrase accurate and aligned with their self-expression. For the two participants who did make edits, the average token-level change was approximately 16\% of the original text, indicating relatively minor revisions.
On average, they found 3.3 out of 5 know-them questions helpful in improving their understanding, and 3.5 out of 5 ask-them questions useful for bringing into a doctor visit.
The expert assessment confirms the clinical accuracy of the outputs. 
The clinical expert rates the faithfulness of the retrieved content at 6.82 out of 7—a very high score. 
The only low rating occurred when a participant who has stage-4 prostate cancer (outside our target group) mistakenly joined the study and received mismatched content.

These findings suggest the system can assist patients in navigating complex medical situations and setting meaningful communication goals for their clinical visits.

\section{Related Work}
Several websites and applications have been developed to aid patient care planning~\cite{mcdarby2021mobile, van2020feasibility}. Most of these tools present web-based forms with limited interactivity. For example, MyDirectives~\cite{mydirectives} uses digital forms combining multiple-choice and open-ended questions to help users document their preferences. 
% The Koda digital ACP platform provides video-based educational content about common life-support treatments and decision-making guidance.
A more recent work, PreCare~\cite{hsu2025precare} explores how AI can support users in clarifying values and improving decision-making confidence.
Despite the promise of existing works, they share several limitations. First, they rarely proactively identify knowledge gaps or provide tailored content to address users’ specific needs. Second, they lack personalization, offering limited support for reflecting individual preferences and personal concerns. Third, many systems focus primarily on value exploration, whereas our system extends beyond reflection to generate actionable outputs—namely, personalized visit questions that patients can both use for self-reflection and take into their clinical appointments.

% \section{Conclusion}

% In this study, we present a system that helps individuals with cancer to prepare their oncology visits.
% To address the knowledge gaps between a uninformed patient and a visit-ready patient. We adapt a famous patient decision aid framework, the Ottawa Personal Decision Guide from it traditional, static format into AI-assisted, RAG-augmented procedure.
% The user study shows Our system can effectively retrieve knowledge related to the perosnal medical concerns, help patients self-reflect among medical options, and recommend visit questions forthe better preparation for their upcoming visits

\section{Conclusion}
In this study, we present a system designed to help individuals with cancer prepare for their oncology visits. 
To bridge the gap between being uninformed and visit-ready, we adapt Ottawa Personal Decision Guide—originally a static patient decision aid—into an AI-assisted, retrieval-augmented generation workflow.
Our system retrieves knowledge tailored to patients' medical concerns, supports self-reflection on treatment options and values, and generates personalized questions to improve communication during the upcoming visit. 
Findings from our user study show that the system is effective, relevant, and easy to use, with strong support for both patient understanding and preparation.

% In this study, we present a system designed to help individuals with cancer prepare for oncology visits.
% To bridge the gap between being uninformed and visit-ready, we adapt the Ottawa Personal Decision Guide—originally a static patient decision aid—into an AI-assisted, retrieval-augmented generation (RAG) workflow.
% Our system retrieves knowledge tailored to each patient's medical concerns, supports self-reflection on treatment options and values, and generates personalized questions to improve communication during the upcoming visit.
% Findings from our user study show that the system is effective, relevant, and easy to use, with strong support for both patient understanding and preparation.

% \begin{table}[ht]
% \centering
% \begin{tabular}{lccc}
% \toprule
% \textbf{Attribute} & \textbf{Mean} & \textbf{Median}  & \textbf{N} \\
% \midrule
% Relevant Score             & 6.70 & 7.00  & 10 \\
% Know-them Count            & 3.30 & 3.50 &  10 \\
% Ask-them Count             & 3.50 & 3.50 &  10 \\
% Faithfulness               & 6.82 & 7 & 10 \\
% \bottomrule
% \end{tabular}
% \caption{Summary statistics for survey attributes (N=10).}
% \label{tab:survey_stats}
% \end{table}

\section*{Limitations}
This study has a few limitations. First, we tested it only on localized prostate cancer, so the results may not apply to other conditions. Second, our user study involved a small number of participants, which may not reflect all patient experiences. Third, the system's personalization depends on the quality of user input—brief answers may lead to less useful outputs. Finally, real-world use would require regular updates to medical content and integration into clinical settings.

% \section*{Acknowledgments}

\bibliography{custom}

\appendix

\begin{figure}[ht]
    \centering
    \includegraphics[width=\linewidth]{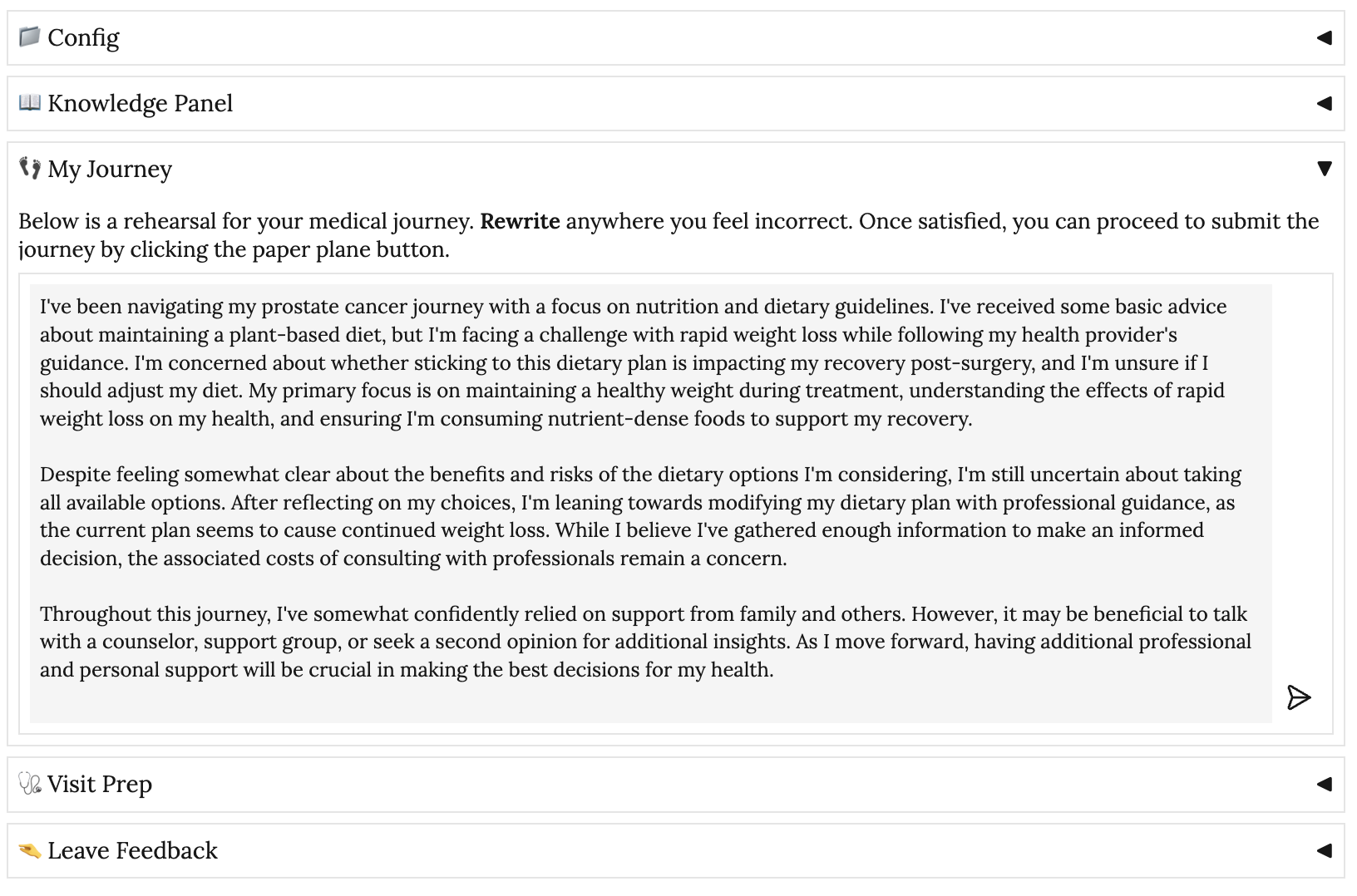}
    \caption{Editable Personal Narrative Section}
    \label{fig:int_journey}
\end{figure}

% \section{Example Appendix}
% \label{sec:appendix}

% This is an appendix.

\end{document}